\documentclass[10pt,letterpaper]{article}
\usepackage[top=0.85in,left=2.75in,footskip=0.75in,marginparwidth=2in]{geometry}

\usepackage[utf8]{inputenc}

\usepackage{cite}

\usepackage{booktabs}
\usepackage{subcaption}
\usepackage{amsmath}

\usepackage{nameref,hyperref}

\usepackage[right]{lineno}

\usepackage{microtype}
\DisableLigatures[f]{encoding = *, family = * }

\raggedright
\usepackage{changepage}

\usepackage[aboveskip=1pt,labelfont=bf,labelsep=period,singlelinecheck=off]{caption}

\makeatletter
\renewcommand{\@biblabel}[1]{\quad#1.}
\makeatother

\usepackage{lastpage,fancyhdr,graphicx}
\usepackage{epstopdf}
\fancyheadoffset[L]{2.25in}
\fancyfootoffset[L]{2.25in}

\usepackage{color}

\definecolor{Gray}{gray}{.25}

\usepackage{graphicx}

\usepackage{sidecap}

\usepackage{wrapfig}
\usepackage[pscoord]{eso-pic}
\usepackage[fulladjust]{marginnote}
\reversemarginpar

\begin{document}
\vspace*{0.35in}

\begin{flushleft}
{\Large
\textbf\newline{Step length measurement in the wild using FMCW radar}
}
\newline
\\
Parthipan Siva \textsuperscript{1,2,*}, 
Alexander Wong \textsuperscript{2}, 
Patricia Hewston\textsuperscript{3,4}, 
George Ioannidis\textsuperscript{3,4}, 
Dr. Jonathan Adachi\textsuperscript{3,4}, 
Dr. Alexander Rabinovich \textsuperscript{5,6}, 
Andrea Lee \textsuperscript{7},
Alexandra Papaioannou \textsuperscript{3,4}
\\
\bigskip
\bf{1} Chirp Inc, Waterloo, ON, Canada
\\
\bf{2} Faculty of Engineering, University of Waterloo, Waterloo, ON N2L 3G1, Canada
\\
\bf{3} Geras Centre for Aging Research, Hamilton Health Sciences, St. Peter’s Hospital, Hamilton, ON
\\
\bf{4} McMaster University, Department of Medicine, Hamilton, ON 
\\
\bf{5} McMaster University, Department of Surgery, Hamilton, ON
\\
\bf{6} ArthroBiologix Inc., Hamilton, ON
\\
\bf{7} Hamilton Health Sciences, Hamilton, ON
\\
\bigskip
* parthipan.siva@mychirp.com

\end{flushleft}

\section*{Abstract}
With an aging population, numerous assistive and monitoring technologies are under development to enable older adults to age in place. To facilitate aging in place predicting risk factors such as falls, and hospitalization and providing early interventions are important. Much of the work on ambient monitoring for risk prediction has centered on gait speed analysis, utilizing privacy-preserving sensors like radar. Despite compelling evidence that monitoring step length, in addition to gait speed, is crucial for predicting risk, radar-based methods have not explored step length measurement in the home. Furthermore, laboratory experiments on step length measurement using radars are limited to proof of concept studies with few healthy subjects. To address this gap, a radar-based step length measurement system for the home is proposed based on detection and tracking using radar point cloud, followed by Doppler speed profiling of the torso to obtain step lengths in the home. The proposed method was evaluated in a clinical environment, involving 35 frail older adults, to establish its validity. Additionally, the method was assessed in people's homes, with 21 frail older adults who had participated in the clinical assessment. The proposed radar-based step length measurement method was compared to the gold standard Zeno Walkway Gait Analysis System, revealing a 4.5cm/8.3\% error in a clinical setting. Furthermore, it exhibited excellent reliability (ICC(2,k)=0.91, 95\% CI 0.82 to 0.96) in uncontrolled home settings. The method also proved accurate in uncontrolled home settings, as indicated by a strong agreement (ICC(3,k)=0.81 (95\% CI 0.53 to 0.92)) between home measurements and in-clinic assessments.


\section{Introduction}
With an aging population, multiple countries are facing challenges caring for older adults. Care facilities are overloaded and hospitals are getting overburdened. Consequently, there has been a shift towards adopting aging-in-place strategies aimed at enabling older adults to stay in their homes for as long as possible while receiving homecare support. Aging in the home is a more scalable solution than building care facilities, and is also the preferable solution for aging individuals. Studies have shown that monitoring aging older adults and individuals with chronic conditions in the home can have a 500\% reduction in cost to the health system \cite{brohman2018}.

To keep people in the home safely, early detection and prediction of frailty, fall risk and hospitalization risk is essential to provide timely interventions and reduce emergency room visits. Gait analysis has been shown to be a predictor of risk factors such as falls, frailty and hospitalization \cite{middleton2015walking,fritz2009white,rodriguez2019spatial,bytycci2021stride,kwon2018comparison}. Gait has many parameters such as speed, step length, cadence, etc. While these parameters are not independent they do have complex relationships. For example, gait speed can be maintained with different step lengths by changing ones cadence.

When someone walks with a shuffling gait -- taking shorter steps, keeping their feet closer to the ground, and leaning forward -- fall risk is increased. Identifying this walking pattern early (by investigating step length) can help in the early detection of falls. In \cite{kraus2022prediction} different gait parameters' importance in predicting frailty was evaluated objectively based on a recursive feature elimination algorithm and ranked by Gini-impurity. Based on the feature importance step length was concluded to be more important than gait speed in predicting frailty \cite{kraus2022prediction}. Furthermore, they note that adding other gait parameters to step length and gait speed had only a slight increase in accuracy. Similarly \cite{woo1999walking} found through a multi-variate analysis that gait speed and step length was important for predicting dependency and mortality but for predicting institutionalization step length alone was the better predictor. For fall risk assessment, \cite{kimura2022step} showed that people with normal gait speed and shorter step length were also at higher risk of falls. \cite{kimura2022step} concluded that gait speed and step length contribute additively to the assessment of fall risk. 

Building upon the insights gleaned from these studies \cite{kraus2022prediction,woo1999walking,kimura2022step,rodriguez2019spatial,bytycci2021stride,kwon2018comparison}, it becomes evident that gait speed and step length play pivotal roles in evaluating frailty, fall risk, and hospitalization risk in older adults. While several methods have been proposed for the continuous monitoring of gait speed in a home environment \cite{atrsaei2021gait,piau2020will,friedrich2022estimating,BETHOUX2018393,joddrell2021continuous,doi:10.1126/scitranslmed.adc9669}, a significant gap exists concerning step length measurement in the home.  

This paper aims to address this critical gap by proposing a radar-based approach for monitoring step length within the home setting. Preliminary approaches, for step length measurement, have been studied in controlled laboratory environments, utilizing cameras \cite{YAGI2020136}, lidar \cite{botros2021contactless}, and radar \cite{abedi2022Hallway}. While camera-based methods are intrusive for in-home use and lidar-based approaches are relatively costly, radar-based solutions offer promise as a privacy-preserving and cost-effective means of measuring step length in a home setting. However, radar has been tested only with young healthy subjects walking 10m or more directly towards the radar. It is essential to ensure that radar-based systems can adapt to the diverse walking patterns of older, frail older adults in a real-world home environment. This paper will investigate the feasibility and effectiveness of radar-based step length measurement for in-home use.

We evaluate the proposed in home step length measurement for reliability using test-retest reliability testing and for validity using correlation with known in-clinic step length measurement. We also present a full in-clinic validation of step length measurement using frail older adults undergoing five different types of walk. This is the first ever evaluation of radar based step length measurement in the clinic and in the home using frail older adults. These in-the-wild studies are needed to validate the use of radar based approaches for continuous in home gait monitoring of older adults aging in place. 

\section{Related Works}

\begin{table}
\renewcommand{\arraystretch}{1.2} 
    \centering
    \begin{small}
    \begin{tabular}{@{}ccccc@{}}
        \toprule
        \textbf{Method} & \textbf{Distance (m)} & \textbf{\# of Participants} & \textbf{Participants} & \textbf{Ground Truth} \\
        \midrule
        \cite{essay97291} & 4 & 3 & Young adults & Marker on shoe \\
        \cite{abedi2023DielectricLense} & 25.2$^\dag$ & 4 & Young adults & Fixed 70 cm steps\\
        \cite{abedi2022Hallway} & 56$^\ddag$ & 5 & Young adults & Fixed 70 cm steps \\
        \cite{sahoTrunkAndToe} & 10 & 10 & Young adults & MOCAP \\
        \bottomrule
    \end{tabular}
    
    $\dag$ 4.2m back and forth three times \hspace{1cm}
    $\ddag$ 14m back and forth two times
    \end{small}
    \caption{Existing radar based step length measurement techniques using trunk movement.}
    \label{tab:existingMethods}
\end{table}

There is a scarcity of controlled setting studies on radar-based step length measurement, with none conducted in an uncontrolled environment. The existing works, outlined in Table \ref{tab:existingMethods}, primarily adopt two approaches: one based on Doppler echoes from the ankle/toes \cite{wangTreadmill,seifertTredmill} and the other on Doppler echoes from the torso \cite{abedi2022Hallway,sahoTrunkAndToe,abedi2023DielectricLense,essay97291}. The ankle/toe-based methods, as highlighted in \cite{sahoTrunkAndToe}, necessitate close proximity of the radar sensor to the walker's feet, making them applicable only in controlled settings, such as treadmill-based studies. Conversely, the torso-based methods are deemed more suitable for ambient step length measurement in a home setting due to the larger size and density of the torso, resulting in stronger radar echoes compared to the ankles/toes.

\begin{figure}
    \centering
    \includegraphics[width=5in]{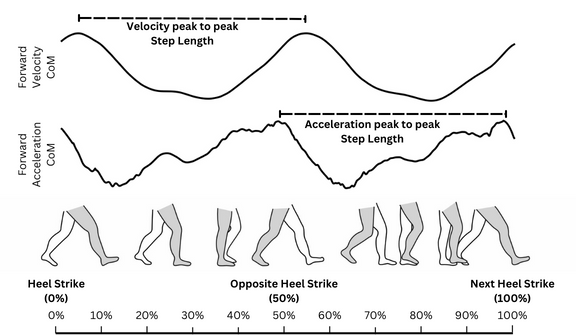}
    \caption{The forward velocity and acceleration of the center of mass during a single gait cycle. The peak to peak distance of velocity and acceleration is equivalent to one step length. Illustration based on speed profile and gait descriptions given in \cite{winter1987biomechanics,s21041264}.}
    \label{fig:gaitCycle}
\end{figure}

The torso-based method hinges on the cyclical pattern of torso speed throughout the gait cycle, as depicted in Figure~\ref{fig:gaitCycle}. Step length is determined by measuring the distance between torso speed peaks \cite{abedi2022Hallway,abedi2023DielectricLense}, the distance between torso acceleration peaks \cite{sahoTrunkAndToe}, or by dividing the average gait speed by the step frequency. The latter is calculated through frequency decomposition of the torso speed profile \cite{essay97291}.

The existing works discussed in Table \ref{tab:existingMethods} exhibit several limitations. Firstly, they predominantly focus on evaluating their methodologies using a limited sample of young, healthy individuals without mobility issues, neglecting the assessment of frail older adults who may exhibit deviations from a healthy gait. Secondly, these methodologies presuppose long, constant-speed walk sequences ranging from 4 to 14 meters or repeated walks up to 56 meters, which proves impractical in a home setting, particularly for older adults who are frail and incapable of maintaining a constant speed over extended distances. Thirdly, the removal of the initial and final 1-2 meters of walk sequences to eliminate acceleration and deceleration effects necessitates even longer walk sequences, thereby excluding the analysis of typical short walks anticipated in a home environment. Lastly, the investigated works do not explore the passive measurement of step length in an unconstrained home environment.

Each individual approach has its own set of limitations. The acceleration peak-to-peak method proposed by \cite{sahoTrunkAndToe} involves taking the derivative of the torso speed profile, making it susceptible to noise inherent in the torso speed measurement. The step frequency-based method introduced by \cite{essay97291} relies on maintaining a constant speed during the walk, achieved by eliminating acceleration and deceleration effects at the walk's start and end. However, this method is impractical for home settings where shorter walk sequences prevent effective compensation for acceleration effects. Consequently, in this study, the torso speed peak-to-peak distance method, as utilized in \cite{abedi2022Hallway, abedi2023DielectricLense}, is employed for step length measurement.

\section{Hardware}

In the proposed approach, the Chirp smart sensor CHIRP-01-T \cite{fcc, chirpwebsite}, affixed to the wall, is employed to monitor individuals, extract torso speeds, and ascertain step length. The Chirp smart sensor is an Internet of Things (IoT) device equipped with onboard processing and utilizes the Texas Instruments IWR6843AOP radar. Due to bandwidth constraints for continuous 24/7 data collection within a home and the limited computational capabilities of the IoT device, only radar point clouds, as detailed in Section~\ref{sec:radarPointCloud}, are processed at a rate of 10 frames per second to track individuals and measure step lengths.

\subsection{Clinical Setup}

\begin{figure}
    \centering
    \includegraphics[width=4.5in]{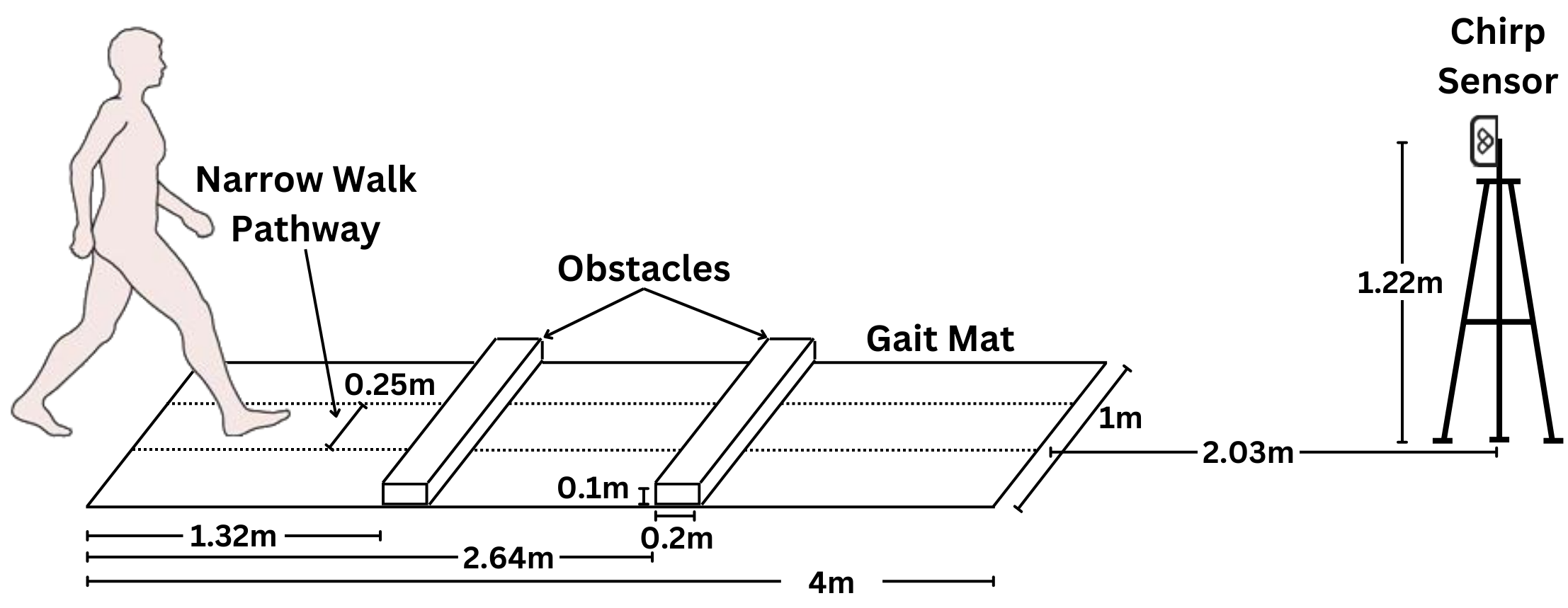}
    \caption{In clinic setup of the 4 meter ProtoKinetics Zeno Walkway Gait Analysis System and Chirp Smart Home Sensor for testing concurrent validity of step length measurement. Obstacles are only used for obstacle walks. Narrow walk pathway is used for narrow walking scenario only.}
    \label{fig:hhsSetup}
\end{figure}

Data collection was conducted in a large multipurpose room within a hospital setting (Figure~\ref{fig:hhsSetup}) with a 4-meter ProtoKinetics Zeno Walkway (Havertown, PA, USA) at a sampling frequency of 100 Hz. The Chirp device (Waterloo, ON, Canada) was positioned at the end of the walking path at a distance of 6.03 meters from the start and 2.03 meters from the end of the ProtoKinetics Walkway at a sampling frequency of 10 Hz (10 frames per second). The ProtoKinetics Zeno Walkway (pressure sensors) and Chirp devices (radar positioning) collected data simultaneously.

The clinical data collection for all participants occurred in multiple sessions over a four-month period. All efforts were made to setup the Zeno Walkway and Chirp sensor at the exact locations specified in Figure~\ref{fig:hhsSetup}. The location of clutter (tables, chairs, etc.) in the room between sessions could vary. During each session 2 to 3 research assistants were present in the room within the field of view of the radar sensor. Furthermore, for older adults with higher frailty severity a research assistant walked behind the individual during their walk across the Zeno Walkway for safety. See Figure~\ref{fig:tracking} for an illustration of a frail older adult walking while two research assistants are nearby for safety.

\subsection{Home Setup}

\begin{figure}
    \begin{subfigure}{.49\textwidth}
        \centering
        \includegraphics[width=0.95\linewidth]{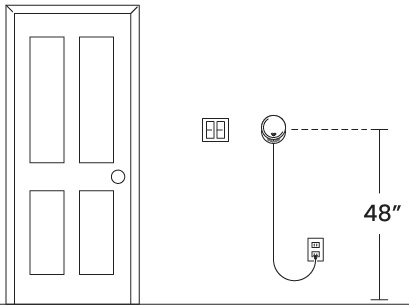}
        \caption{Elevation at switch height.}
        \label{fig:wallHeight}
    \end{subfigure}
    \begin{subfigure}{.49\textwidth}
        \centering
        \includegraphics[width=0.95\linewidth]{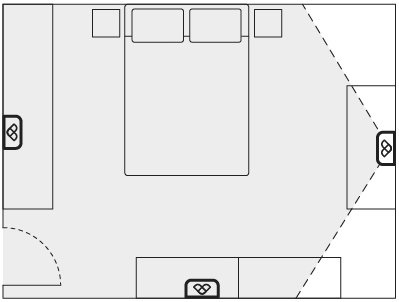}
        \caption{Possible locations of the Chirp sensor at center of wall covering entire room.}
        \label{fig:inRoom}
    \end{subfigure}
    \caption{Placement of Chirp sensor in the room.}
    \label{fig:chirpSetup}
\end{figure}

For in-home step length measurement, participants were directed to install Chirp sensors in their bedroom, living room, and kitchen. Guidelines were provided to position the Chirp sensor between 121 cm (48 inches) and 132 cm (52 inches) above the floor, which corresponds to the typical height of residential wall switches (see Figure \ref{fig:wallHeight}). Participants were further instructed to place the Chirp sensor as centrally as possible on the wall, ensuring full coverage of the room (see Figure \ref{fig:inRoom}). Following installation, participants utilized the Chirp Labs App to connect the Chirp sensor to their Wi-Fi and assigned names as bedroom, kitchen, and living room to each respective sensor.

For inter-device test-retest reliability within the home between week 1 and week 2, participants were requested to remount all Chirp sensor devices after the first week of data collection (e.g., relocating the kitchen device to the bedroom, the bedroom device to the living room, and the living room device to the kitchen).

The installation and setup process was left to the discretion of the user, and the authors did not modify or validate the device placement. Consequently, the placement reflects how families might set up the devices in a consumer setting.

\section{Proposed Approach}

\begin{figure}
    \centering
    \includegraphics[width=5.2in]{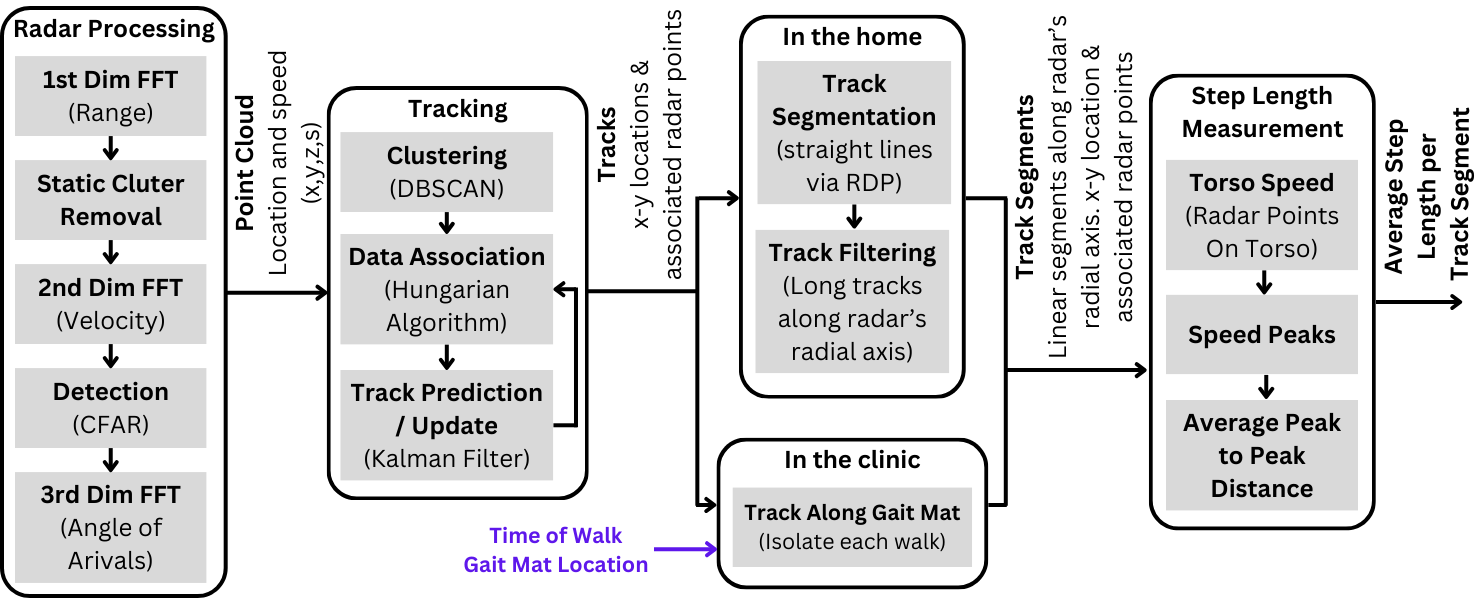}
    \caption{Step length measurement methodology: Radar signal processing generates 3D point clouds with speeds, enabling detection and tracking of individuals. In the home, linear track segments along the radar's radial axis are isolated, while in the clinic, track segments along Zeno Walkway's linear path are extracted. Step length is determined as the peak-to-peak distance of torso speed.}
    \label{fig:algOverview}
\end{figure}

The overall approach for step length measurement in a home setting and clinical setting is illustrated in Figure~\ref{fig:algOverview}. The TI signal processing SDK \cite{timmwavesdk} is used to produce a radar point clouds at 10 frames per second (fps). The point clouds are used to detect and track individuals moving in the scene. The tracks are then filtered to identify tracks that can be used for step length measurement in the home or in the clinical setting. Once viable tracks are identified, the torso speed along the track is used to measure the average step length.

\subsection{Radar Point Cloud}
\label{sec:radarPointCloud}

The Texas Instrument radar processing tool chain \cite{timmwavesdk} consisting of signal processing, static clutter removal and constant false alarm rate (CFAR) detection is used to obtain a radar point cloud at time $t$. Point cloud is formed by a set of moving points detected by the radar (Figure~\ref{fig:radarPoints}), where each point consists of a location and speed.

\begin{equation}
    \mathcal{P}_t = \{p_t^1,\ldots,p_t^i,\ldots,p_t^n\} \label{eq:radarCloud}
\end{equation}

\noindent where $t$ is the current time, and the $i^{th}$ point is $p_t^i = (x_i,y_i,z_i,s_i)$. Location $x,y,z$ is in meters and speed $s$ is in meters per second.

\subsection{Detection and Tracking}

\begin{figure}
    \begin{subfigure}{.32\textwidth}
        \centering
        \includegraphics[width=0.98\linewidth]{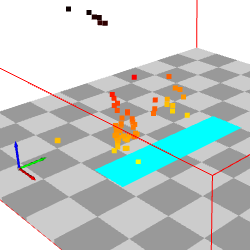}
        \caption{Radar point cloud.}
        \label{fig:radarPoints}
    \end{subfigure}
    \begin{subfigure}{.32\textwidth}
        \centering
        \includegraphics[width=0.98\linewidth]{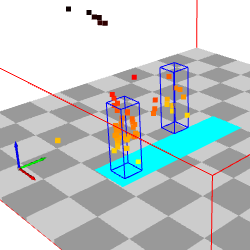}
        \caption{Detection (DBSCAN).}
        \label{fig:detection}
    \end{subfigure}
    \begin{subfigure}{.32\textwidth}
        \centering
        \includegraphics[width=0.98\linewidth]{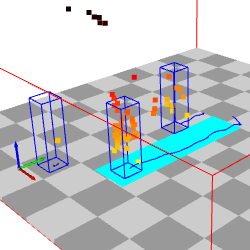}
        \caption{Tracking (Kalman Filter).}
        \label{fig:tracking}
    \end{subfigure}
    \caption{Tracking illustration: Radar point clouds are clustered to form detections, which are associated to tracks through the Hungarian Algorithm, and tracked using Kalman Filtering. The floor is depicted with a 1m by 1m checkerboard pattern, while Zeno Walkway is represented by a cyan rectangle.}
    \label{fig:detectionTracking}
\end{figure}

Using the radar point cloud $\mathcal{P}_t$ a detection and tracking approach based on DBSCAN clustering for detection (Figure~\ref{fig:detection}), data association via Hungarian assignment and Kalman Filtering for tracking (Figure~\ref{fig:tracking}) as outlined in \cite{radarTracking2019Zhao} is used for tracking people in the scene.

This results in a set of tracks:

\begin{equation}
    \mathcal{T} = \{T_1,\ldots,T_i,\ldots,T_N\} \label{eq:trackSet}
\end{equation}

\noindent Each track is defined as:

\begin{equation}
    T_i = \{(x_{t_0},y_{t_0},\mathcal{S}_{t_0}),\ldots,(x_{t_j},y_{t_j},\mathcal{S}_{t_j}),\ldots,(x_{t_N},y_{t_N},\mathcal{S}_{t_N})\}
\end{equation}

\noindent where $\mathcal{S}_{t_j} \subseteq \mathcal{P}_{t_j}$ ($\mathcal{P}$ is defined in \eqref{eq:radarCloud}) and $(x_{t_j},y_{t_j})$ is the track location in the room at time $t_j$. As in \cite{radarTracking2019Zhao} we track moving objects' location only in the $x-y$ plane, ignoring elevation $z$. $\mathcal{S}_{t_j}$ is formed by DBSCAN clustering from radar point cloud $\mathcal{P}_{t_j}$ at time $t_j$ and assigned to track $T_i$ during data association via Hungarian assignment.

\subsection{Tracks in the Clinic}

In the clinic, the radar is setup in front of a Zeno Walkway (Figure~\ref{fig:hhsSetup}). For fair comparison, radar based step length measurement must be conducted over the track segment starting and ending on the Zeno Walkway. Clinic setting is reproduced for each participant such that the Zeno Walkway start and ends at coordinates $g_s = (0, 6.03)$ and $g_e = (0, 2.03)$, respectively. Furthermore, for each walk $w$ by participant $i$, a start time $t_i^w$ and Zeno Walkway average step length $g_i^w$ was recorded. 

Given all the tracks $\mathcal{T}$ (defined in \eqref{eq:trackSet}) obtained during participant testing, the track segment associated with the in-clinic walks are

\begin{equation}
    \mathcal{L}' = \{L_1^1,L_1^2,\ldots,L_1^W,\ldots,L_i^k,\ldots,L_P^W\} \label{eq:clinicLineSegOfInterest}
\end{equation}

\noindent where $P$ is the number of participants and $W$ is the number of walks for each participant. The track segment $L_i^k$ is obtained as the track segment starting near $g_s$ and ending near $g_e$ and is closest in starting time to $t_i^w$.

\subsection{Tracks in the Home}

The Doppler radar is most sensitive to speed changes along the radial axis. As a result, the best way to isolate small fluctuations in torso speed, which are caused during the normal gait cycle, is to look at the torso speed when a person is travelling along the radial axis of the radar. To this end, given all the tracks $\mathcal{T}$ from a home setting, the track segments that are relatively a straight line going along the radial axis is isolated for step length measurement. First, all tracks are segmented into linear segments (Section~\ref{sec:trackSegmentation}) and then the linear segments are classified as valid segments traveling along the radar's radial axis (Section~\ref{sec:trackSegmentClassification}).

\subsubsection{Track Segmentation}
\label{sec:trackSegmentation}

To isolate instances where individuals are walking towards or away from the radar (i.e., along radar's radial axis), we segment all tracks into linear segments. The $x-y$ locations of the track $T_i$ is treated as a polyline, which is decimated using the Ramer–Douglas–Peucker (RDP) algorithm \cite{hershberger1994n}. The RDP algorithm has a single parameter $\varepsilon$ which controls the decimation and is the maximal distance allowed between a point on the polyline and the linear representation of that polyline.

\begin{figure}
    \begin{minipage}{0.47\textwidth}
        \includegraphics[width=\linewidth]{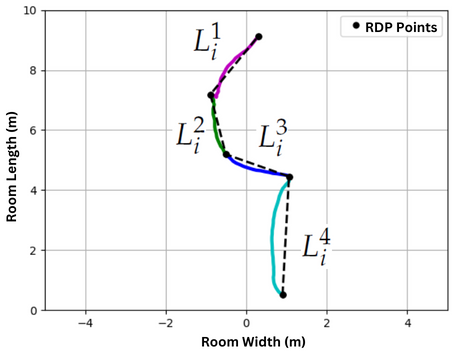}
        \caption{Track $T_i$ is segmented into linear track segments $L_i^1,\ldots,L_i^4$ using the Ramer-Douglas-Peucker algorithm. $\varepsilon=0.5m$ for this figure.}
        \label{fig:RDP}
    \end{minipage}
    \hspace{0.2cm}
    \begin{minipage}{0.47\textwidth}
        \includegraphics[width=\linewidth]{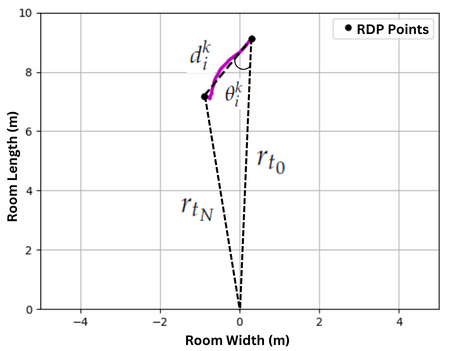}
        \caption{Linear track segment $L_i^k$ has a length $d_i^k$ \eqref{eq:Ldist} and an orientation $\theta_i^k$ \eqref{eq:Lori} that is needed to orient the track along radar's radial axis.}
        \label{fig:Ori}
    \end{minipage}
\end{figure}

The points selected by the RDP decimation is used to segment track $T_i$ into linear track segments. If RDP algorithms selects $M_i+1$ points along track $T_i$ to keep, then track $T_i$ will be segmented into $M_i$ linear segments as illustrated in Figure~\ref{fig:RDP}. We represent the linear segments as:
f
\begin{equation}
    T_i = \{L_i^1,\ldots,L_i^k,\ldots,L_i^{M_i}\}
\end{equation}

The set of all linear track segments becomes:

\begin{equation}
    \mathcal{L} = \{L_1^1,\ldots,L_1^{M_1},\ldots,L_i^1,\ldots,L_i^k,\ldots,L_i^{M_i},\ldots,L_N^{M_N}\}
\end{equation}

\noindent where $N$ is the number of tracks as defined in \eqref{eq:trackSet}, $M_i$ is the number of linear segments in track $i$ and the cardinality of the set, $|\mathcal{L}| = \sum_{i=1}^{N} M_i = m$, is the number of linear track segments from the home.

\subsubsection{Track Segment Classification}
\label{sec:trackSegmentClassification}

Given linear track segments, the segments along the radial axis of the radar must be isolated. To this end, given a linear track segment

\begin{equation}
L_i^k = \{(x_{t_0},y_{t_0},\mathcal{S}_{t_0}),\ldots,(x_{t_j},y_{t_j},\mathcal{S}_{t_j}),\ldots,(x_{t_N},y_{t_N},\mathcal{S}_{t_N})\}
\end{equation}

\noindent we define

\begin{align}
    r_{t_0} &= \sqrt{(x_{t_0})^2+(y_{t_0})^2} \label{eq:rToStart} \\
    r_{t_N} &= \sqrt{(x_{t_N})^2+(y_{t_N})^2} \label{eq:rToEnd}\\
    d_i^k &= \sqrt{(x_{t_0}-x_{t_N})^2+(y_{t_0}-y_{t_N})^2} \label{eq:Ldist} \\ 
    \theta_i^k &= 
\arccos{\left( \frac{(\max(r_{t_0},r_{t_N}))^2 + (d_i^k)^2 - (\min(r_{t_0},r_{t_N}))^2}{2d_i^k\max(r_{t_0},r_{t_N})} \right)} \label{eq:Lori}
\end{align}

\noindent As illustrated in Figure~\ref{fig:Ori}, $\theta_i^k$ is the rotation angle needed, about the radially furthest track endpoint, to rotate the track directly towards (away from) the radar. The classification of track segment $L_i^k$ is:

\begin{equation}
    c_i^k = \begin{cases}
        1 & \text{if } d_i^k \geq D \text{ and } \theta_i^k \leq \gamma \\
        0 & \text{otherwise}
    \end{cases}
    \label{eq:classificationGoodTrack}
\end{equation}

\noindent where $c_i^k=1$ indicates valid linear tracks segment along radar's radial axis that is sufficiently long enough to detect step lengths and its relative orientation to radar's radial axis is small.

This results in a set of valid radially aligned linear track segments:

\begin{equation}
    \mathcal{L}' = \{L_i^k \in \mathcal{L} : c_i^k = 1\} \label{eq:homeLineSegsOfInterest}
\end{equation}

\subsection{Step Length Measurement}

Given a linear track segment along radar's radial axis, $L_i^k \in \mathcal{L}'$, from the home \eqref{eq:homeLineSegsOfInterest} or at the clinic \eqref{eq:clinicLineSegOfInterest}, the average step length needs to be measured. Similar to \cite{abedi2022Hallway,abedi2023DielectricLense}, step length measurement is obtained as the peak to peak distance of the torso speed.

\subsubsection{Torso Speed}

Each linear track segment $L_i^k \in \mathcal{L}'$ has a set of tracked locations:

\begin{equation}
\label{eq:linearTrackSegment}
L_i^k = \{(x_{t_0},y_{t_0},\mathcal{S}_{t_0}),\ldots,(x_{t_j},y_{t_j},\mathcal{S}_{t_j}),\ldots,(x_{t_N},y_{t_N},\mathcal{S}_{t_N})\}
\end{equation}

\noindent where 

\begin{equation}
    \mathcal{S}_{t_j} = \{(x_1,y_1,z_1,s_1),\ldots,(x_a,y_a,z_a,s_a),\ldots,(x_N,y_N,z_N,s_N)\}
\end{equation}

\noindent represents the set of radar points on the person being tracked, which includes points on the torso, arms, legs etc. From this, points on the torso $\mathcal{S}_{t_j}^{torso} \in \mathcal{S}_{t_j}$ is isolated based on elevation data and direction of travel.

The radar is placed 121 cm (48 inches) to 132 cm (52 inches) above the floor as such we conservatively estimate torso points to be in the range $-Z_{torso} \leq z_a \leq Z_{torso}$. Furthermore, if the person is travelling towards the radar we expect the Doppler speed of the torso to be $-ve$ and if the person is travelling away from the radar we expect the Doppler speed of the torso to be $+ve$. This distinction is important because the arms can be travelling in the opposite direction to the torso.

Specifically, given linear track segment $L_i^k$ in \eqref{eq:linearTrackSegment}, the radial distance to the start $r_{t_0}$ \eqref{eq:rToStart} and end $r_{t_N}$ \eqref{eq:rToEnd} locations, the radar points on the torso $\mathcal{S}^{torso}_{t_j} \subseteq \mathcal{S}_{t_j}$ is defined as

\begin{align}
\mathcal{S}^{torso}_{t_j} &= \{(x_a,y_a,z_a,s_a) \in \mathcal{S}_{t_j} : -Z_{torso} \leq z_a \leq Z_{torso} \text{ and } \alpha s_a > 0 \} \label{eq:torsopts}\\
\alpha &= \begin{cases}
        -1 & \text{if } r_{t_N} < r_{t_0}\\
        +1 & \text{if } r_{t_N} > r_{t_0}
    \end{cases}
\end{align}

\noindent The torso speed at each time step $t_j$ is $v_{t_j}$ and is computed as the average of speeds in $\mathcal{S}^{torso}_{t_j}$.

\begin{equation}
    v_{t_j} = \frac{\sum_a s_a \in \mathcal{S}^{torso}_{t_j}}{|\mathcal{S}'_{t_j}|}
\end{equation}

Given a set of torso speed $v_{t_j}$ the linear track segment $L_i^k$ becomes:

\begin{equation}
\label{eq:linearTrackSegmentVelocity}
L_i^k = \{(x_{t_0},y_{t_0},v_{t_0}),\ldots,(x_{t_j},y_{t_j},v_{t_j}),\ldots,(x_{t_N},y_{t_N},v_{t_N})\}
\end{equation}

\noindent where $x,y$ is the location of the person and $v$ is the Doppler torso speed of the person.

Acquiring torso speed from radar points linked to a track offers a benefit. In situations with multiple individuals within the radar field of view, as depicted in Figure~\ref{fig:tracking}, it enables the determination of accurate torso speeds for each person. This enhances the reliability of torso speed estimation, particularly in noisy conditions.

\subsubsection{Peak to Peak Distance}
\label{sec:peakToPeakDist}

\begin{figure}
    \centering
    \includegraphics[width=.9\linewidth]{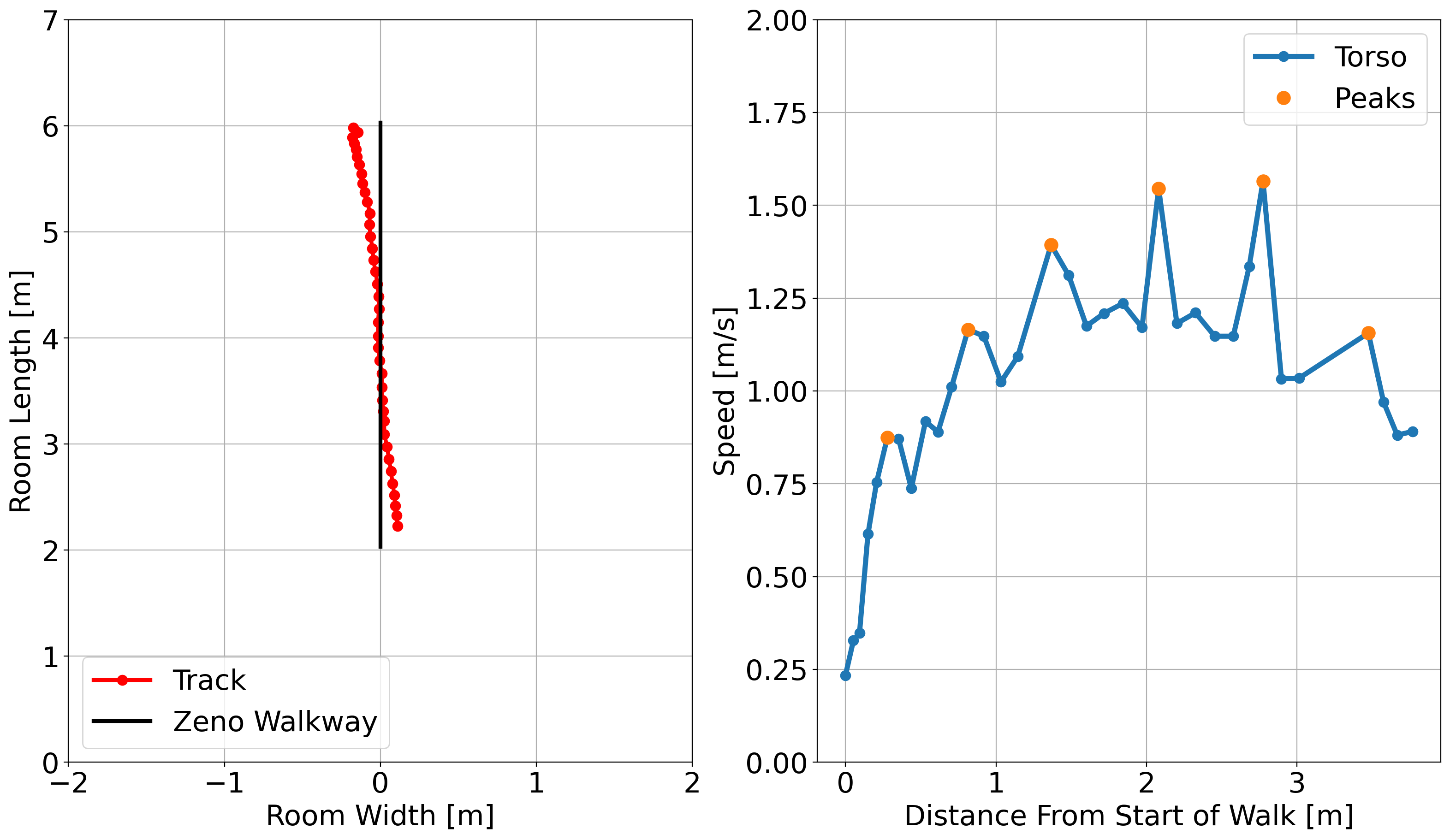}
    \caption{Control (Normal speed) walk by a participant: On the left, the tracked location overlaid with the expected location of the Zeno Walkway. On the right, the Doppler torso speed, featuring detected torso speed spikes. Notably, the speed trend is non-constant due to the acceleration and deceleration effects of starting and stopping.}
    \label{fig:samplePeaks}
\end{figure}

The torso speeds from \eqref{eq:linearTrackSegmentVelocity} is used to find peaks as illustrated in Figure~\ref{fig:samplePeaks}. Given the torso speeds on the linear track segment, a center surrounded window of $0.4sec$ (5 points given a 10FPS) is used for non-maximum-suppression (NMS). After NMS, all peaks are found and sorted in descending order of speed. Starting at the fastest speed (highest peak), peaks are kept as valid peak if the peak to peak time is at least $R$ seconds. Once all valid peaks are found, the peak to peak distance and peak to peak time is obtained as potential step length and step time.

The peak detection algorithm may skip a step due to noise or the irregular gait of frail older adults, leading to inaccuracies in measuring peak-to-peak distances and resulting in larger step lengths. To account for potential missed steps, any step lengths exceeding $1m$ or step times exceeding $3$ seconds are excluded as outliers. Subsequently, if the linear track segment has a minimum of two measured step lengths, the average of these step lengths is calculated and considered as the average step length for the linear line segment $L_i^k$.

\section{Experiment Setup}

The study included individuals aged 60 years and older who met the following inclusion criteria: (1) frailty, indicated by a score of 3 or more on the FRAIL Scale, (2) lived alone, (3) had a home Wi-Fi connection and (4) access to a smartphone or tablet for device setup. Exclusion criteria encompassed individuals who required a wheelchair for indoor mobility, needed prolonged sitting due to a medical condition, or lacked independent mobility. Participants with travel plans or commitments missing more than 30\% of the study period were excluded. Participants were recruited from regional specialized geriatric clinics, community groups, and newspaper advertisements. This study was approved by Hamilton Integrated Research Ethics Board (HIREB Project\# 15237) and was performed in accordance with the Declaration of Helsinki. Written informed consent was obtained from all participants.

\begin{table}
\renewcommand{\arraystretch}{1.2} 
    \centering
    \begin{small}
    \begin{tabular}{@{}ll@{}}
        \toprule
        \textbf{Demographics}&\textbf{All Participants (N=35)}\\
        \midrule
        Age, M (SD)&75.49 (6.56)\\
        Age, Range& 60 to 89\\
        Sex, \% female &30/35 (85.71\%)\\
        Education, n more than high school&23/35 (65.71\%)\\
        Living arrangement, n lives alone&35/35 (100\%)\\
        Physical function, SPPB total score, M (SD)&8.53 (2.74)\\
        Physical function, n SPPB <9&12/34 (35.29\%)\\
        Fear of falling, FES-I total score, M (SD)&24.97 (6.62)\\
        Fear of falling, n FES-I moderate to high severity &26/34 (76.47\%)\\
        Cognition, MOCA total score,  M (SD)&23.38 (3.64)\\
        Cognition, n MOCA total score <25&20/34 (58.82\%)\\
        \bottomrule
        
\end{tabular}
\end{small}
\caption{Demographics of the participants.}
\label{tab:demographic}
\end{table}

Participants' demographic information (Table~\ref{tab:demographic}) was collected of age, sex and education. Physical performance was assessed with the Short Physical Performance Battery (SPPB) \cite{10.1093/geronj/49.2.M85}. An SPPB score of <9 points indicates poor physical performance and is predictive of hospitalization and mortality \cite{Negm__2019}. The Falls-Efficacy Scale International (FES-I) is a standardized questionnaire that assesses concerns about falling within 16 physical and social activities at home and the community. FES-I items are rated on a four-point scale (1 [not at all concerned] to 4 [very concerned]) and total scores range from 16-64. FES-I total scores can be further classified based on fear of falling severity with clinical cut-off points of no to low (score=16-19) and moderate to high (score=20-64) concern about falling \cite{yardley2005development}. Cognition was assessed with the Montreal Cognitive Assessment (MoCA) \cite{nasreddine2005montreal}. Total MOCA scores range from 0-30, and >26 points are considered normal cognitive function \cite{nasreddine2005montreal}.

\subsection{Clinic Setup}

Using the InCIANTI protocol \cite{deshpande2013can}, participants walked along the 4-meter path during normal [control] and adaptative locomotion experimental conditions (walking-while-talking [dual task] reciting animal names from a given letter; obstacle crossing of two 4.5 inch high obstacles; narrow walking along a 25cm wide path; fast walking). Each of the 5 experimental condition was conducted twice in a randomized order except for fast walking trials which were consistently performed last in each experimental block to avoid any influence on the speed of the preceding trials. Two blocks of walks was conducted within a participant session separated by approximately 30 minutes for intra-session reliability testing.

Participants commenced their walks at the beginning of the Zeno Walkway, proceeded to the end, and then came to a stop. Notably, unlike existing works \cite{essay97291,abedi2022Hallway,abedi2023DielectricLense,sahoTrunkAndToe}, this approach encompasses the acceleration and deceleration effects associated with walking. This methodology aligns with testing in home environments, as the limited space within homes makes it unfeasible to omit the acceleration and deceleration segments of the walks.

\begin{table}
\renewcommand{\arraystretch}{1.2} 
    \centering
    \begin{small}
    \begin{tabular}{@{}lcccccc@{}}
        \toprule
        &\textbf{Control}&\textbf{Fast}&\textbf{Narrow}&\textbf{Obstacle}&\textbf{Dual Task}&\textbf{All}\\
        \midrule
        \textbf{Tech. Difficulty}&6&4&6&5&7&28 (4.0\%)\\
        \textbf{Unable}&1&4&17&24&1&47 (6.7\%)\\
        \midrule
        \textbf{Collected Walks}&133&132&117&111&132&625 (89.3\%)\\
        \bottomrule
\end{tabular}
\end{small}
\caption{Of the 700 walks (35 participant $\times$ 20 walks), data is missed due to technical difficulties, and participants unable to complete the walks due to frailty.}
\label{tab:dataProblems}
\end{table}

\begin{figure}
    \centering
    \includegraphics[width=\linewidth]{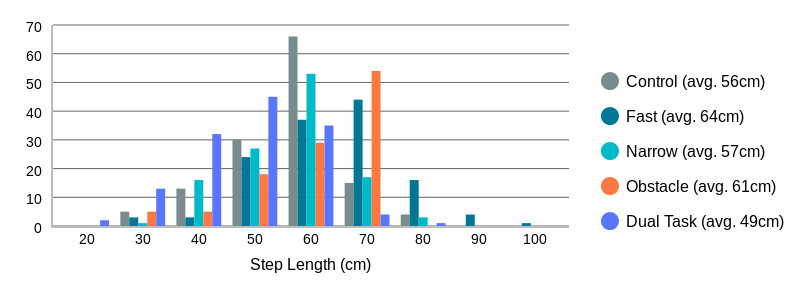}
    \caption{Ground truth step length distribution as measured by the Zeno Walkway Gait Analysis System. Overall average step length is 57cm (12cm).}
    \label{fig:steplengthdist}
\end{figure}

As shown in Table~\ref{tab:dataProblems}, of the 700 walks (35 participants $\times$ 2 blocks $\times$ 2 repetition $\times$ 5 types of walks), 47 walks (across 6 participants) were not collected because the participants was too frail to complete the walks. A further 28 walks (across 4 participants) were not collected due to technical issues. Remaining 625 walks were collected with Chirp sensor and Zeno Walkway. The collected step lengths has a robust variation with a mean of 57cm (12cm) as shown in Figure~\ref{fig:steplengthdist}.

\subsection{Home}

Participants installed the devices in their living room, bedroom, and kitchen for a two-week duration. Out of the 35 participants, 21 successfully set up the devices in their homes. Among these 21 participants, 3 have data for only one week, 1 participant has data for 10 days, and the remaining 17 have data for the full two weeks.

All participants reside alone. To ensure that the in-home evaluation of step length pertains solely to the participant, step length is reported only when a single person is being tracked within the home. Some participants have small pets, such as cats, which are excluded from tracking based on their size.

\section{Algorithm Parameters}

In home track selection for step length measurement is dependent on three parameters: RDP threshold $\varepsilon$ (Section \ref{sec:trackSegmentClassification}), length of linear track segment $D$ \eqref{eq:classificationGoodTrack} and orientation threshold $\gamma$ \eqref{eq:classificationGoodTrack}. Both $\varepsilon$ and $\gamma$ are set empirically as $\varepsilon=0.5m$ and $\gamma=15^o$. Minimum length of track $D$ must be selected to ensure at least two step lengths are present within the linear track segment (Section~\ref{sec:peakToPeakDist}). Per \cite{zarrugh1974optimization}, the average male step length for a walking speed of 1.6 m/s is 0.84m. This requires a track length of at least 1.7m for two steps. Based on this upper bound, $D$ is set as $D=2m$.

The torso location cutoff, as defined in \eqref{eq:torsopts}, must be defined based on known radar configuration. Participants were given instructions to setup the radar approximately at a height of 121cm. Based on that $Z_{torso}$ is conservatively set as $Z_{torso} = 0.25m$. This assumes that the torso radar points are within 0.25m below to 0.25m above the radar. This is empirically set based on male average torso length of 46cm to 52cm.

Finally the minimum peak to peak time $R$ (Section~\ref{sec:peakToPeakDist}) is set as small as possible. Given 10fps radar data and center surrounded non-maximum suppression window of 0.4sec (Section~\ref{sec:peakToPeakDist}), the lower bound on $R$ is $R>0.4/2=0.2sec$. Based on this, the minimum peak to peak time is set as $R=0.3sec$.

\section{In Clinic Step Length Evaluation}

The study gathered data from 625 walks conducted in a clinic, with step length measurements simultaneously obtained from both the Chirp sensor and Zeno Walkway. This dataset serves as the basis for several evaluations. Firstly, an analysis of step length detection rate is conducted using the proposed method. Secondly, the concurrent validity of the proposed method is assessed against the gold standard measurement obtained from the Zeno Walkway. Finally, a comparison of the proposed method with existing methods is undertaken.

\subsection{Step Length Detection Rate}

\begin{table}
\renewcommand{\arraystretch}{1.2} 
    \centering
    \begin{small}
    \begin{tabular}{@{}lcccccc@{}}
        \toprule
        &\textbf{Control}&\textbf{Fast}&\textbf{Narrow}&\textbf{Obstacle}&\textbf{DualTask}&\textbf{All}\\
        \midrule
        \textbf{Alg. Missed}&2&18&6&0&0&26/625 (4.2\%)\\
        \textbf{Alg. Detected}&131&114&111&111&132&599/625 (95.8\%)\\
        \bottomrule
\end{tabular}
\end{small}
\caption{Step length detection rates for the proposed approach on the 625 walks.}
\label{tab:detectionRate}
\end{table}

Over the 4-meter walk, it is required to detect at least two step length measurements (i.e., three consecutive torso speed peaks) to generate an average step length measurement for the walk (refer to Section~\ref{sec:peakToPeakDist}). Consequently, there are instances where the proposed method does not provide a step length measurement. Out of the 625 walks, the proposed method successfully detected step length in 599 walks, yielding a detection rate of 96\%. The distribution of missed walk types is detailed in Table~\ref{tab:detectionRate}.

\begin{figure}
    \centering
    \includegraphics[width=0.95\linewidth]{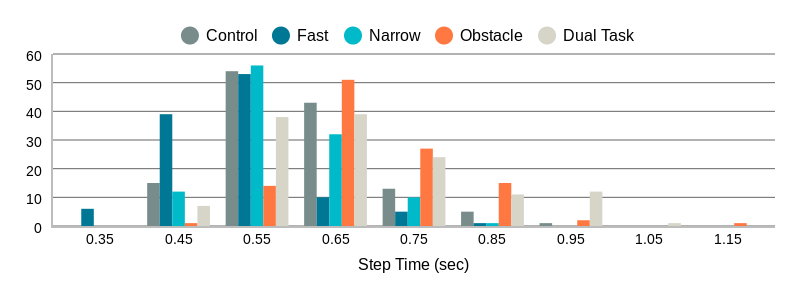}
    \caption{Distribution of step times as measured by torso speed peak to peak times.}
    \label{fig:stepTimeDist}
\end{figure}

As indicated in Table~\ref{tab:detectionRate}, the majority of missed step length measurements are associated with fast walks. This can be attributed primarily to the set minimum peak-to-peak time threshold of $R=0.3$ seconds. The distribution of detected step length peak-to-peak times (i.e., step times) is illustrated in Figure~\ref{fig:stepTimeDist}. Notably, the distribution of step times for fast walks crosses the $R=0.3$ second threshold. Consequently, the peak-to-peak measure for fast walks becomes undetectable in certain cases. To enable step length measurement at higher walking speeds, generating radar point clouds at a higher frame rate than 10 FPS and lowering the minimum peak-to-peak distance threshold from $R=0.3$ seconds is necessary.

In the context of in-home monitoring, there is no need for adjustments to the radar frame rate and peak-to-peak distance threshold. The step length measured at home typically appears smaller than what is observed in the clinic, as discussed in Section~\ref{sec:inhomevaldity}.

\subsection{Concurrent Validity}

\begin{figure}
    \begin{minipage}{0.48\textwidth}
        \includegraphics[width=\linewidth]{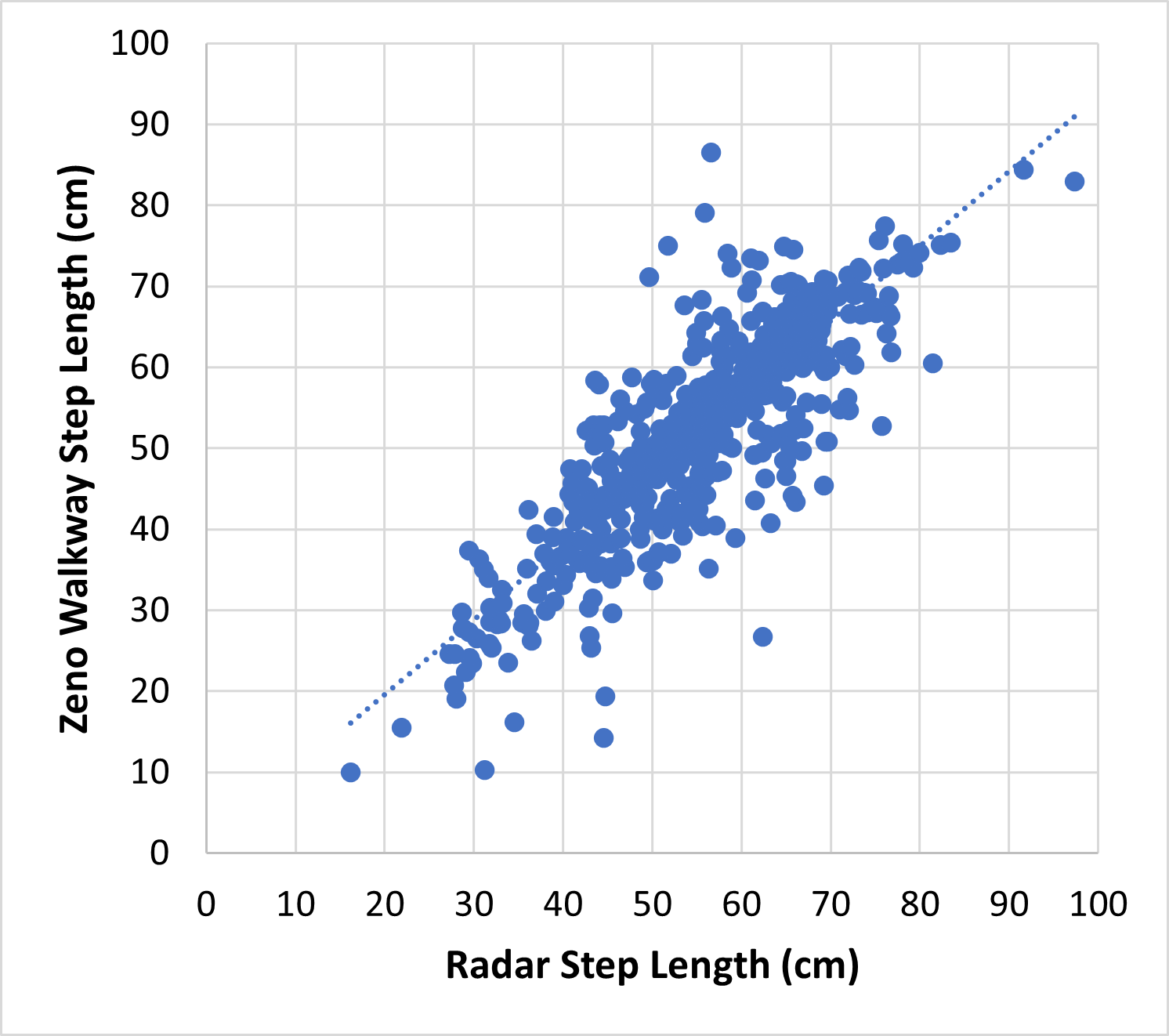}
        \caption{Proposed radar-based step length measurement vs. Zeno Walkway. Data includes all 599 walks (4-meter each) by 35 frail older adults.}
        \label{fig:allWalks}
    \end{minipage}
    \hspace{0.2cm}
    \begin{minipage}{0.48\textwidth}
        \includegraphics[width=\linewidth]{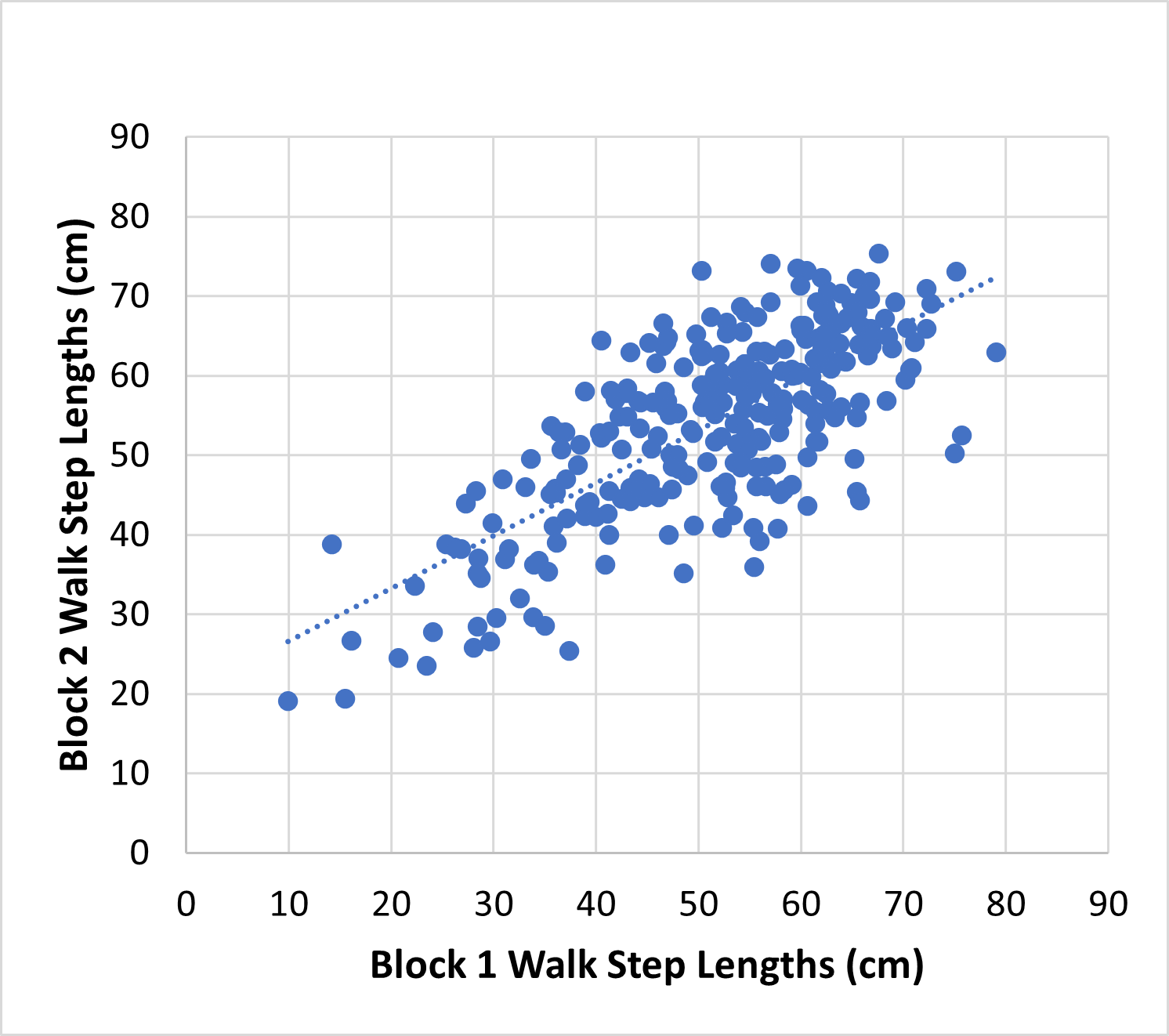}
        \caption{Intra-session reliability. Block 1 to Block 2 step lengths measured in the clinic. ICC(2,k) = 0.83 (95\% CI 0.77 to 0.87)}
        \label{fig:intrasessionreliability}
    \end{minipage}
\end{figure}

\begin{figure}
\begin{adjustwidth}{-1.5in}{0in}
    \begin{subfigure}{.64\textwidth}
        \centering
        \includegraphics[width=0.98\linewidth]{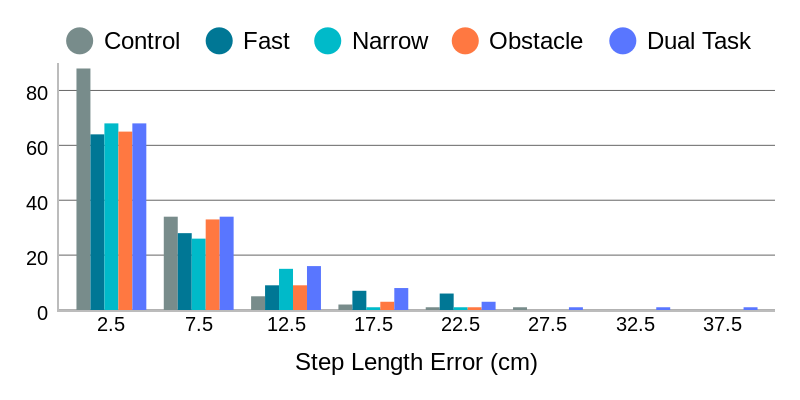}
        \caption{Error distribution in cm.}
        \label{fig:steplenerrordistcm}
    \end{subfigure}
    \begin{subfigure}{.64\textwidth}
        \centering
        \includegraphics[width=0.98\linewidth]{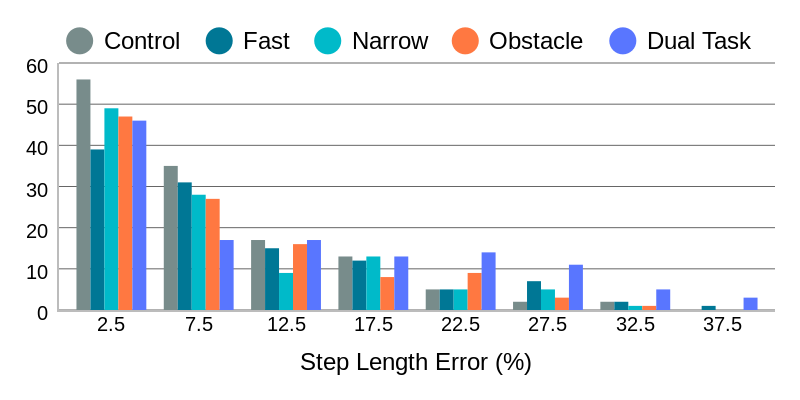}
        \caption{Error distribution \% of step length.}
        \label{fig:steplenerrordistpct}
    \end{subfigure}
    \caption{Step length error distribution by walk type in cm and as a percentage of the true step length measured by Zeno Walkway.}
    \label{fig:steplenerrordist}
\end{adjustwidth}
\end{figure}

Figure~\ref{fig:allWalks} displays all step length measurements obtained through the proposed radar-based method in comparison to those from the Zeno Walkway. Concurrent validity is evaluated by considering the absolute difference between Zeno Walkway step length measurements and those obtained using the proposed radar-based method. The analysis involves a total of 599 walks, where the proposed algorithm reported a step length. To address variations in individuals' step lengths (ranging from 26cm to 97cm, as illustrated in Figure~\ref{fig:steplengthdist}), step length errors are further expressed as a percentage of the Zeno Walkway step length measurement. The comprehensive step length errors, along with a breakdown by different walk types, are presented in Table~\ref{tab:steplenerror} and Figure~\ref{fig:steplenerrordist}.

In the control walk (i.e., normal walking speed), the average error is 4.5cm, which, on average, is less than 10\% of the true step length. Both narrow walkway walks and obstacle walks exhibit comparable absolute and relative errors. Fast walks and dual task walks share the same absolute average error of 6.5cm. However, the relative error for the dual task is 4\% higher. This discrepancy arises from the fact that dual task walks have shorter step lengths (refer to Figure~\ref{fig:steplengthdist}) compared to fast walks. This underscores the significance of considering both absolute and relative step length errors in the assessment of step length measurement methods.

\begin{table}
\begin{adjustwidth}{-1.5in}{0cm}
\renewcommand{\arraystretch}{1.2} 
    \centering
    \begin{small}
    \begin{tabular}{@{}lcccccc@{}}
        \toprule
&\textbf{Control}&\textbf{Fast}&\textbf{Narrow}&\textbf{Obstacle}&\textbf{Dual Task}&\textbf{All}\\
\midrule
\textbf{cm} &4.5 (4.3)&6.5 (5.9)&5.0 (4.3)&5.0 (4.4)&6.5 (6.4)&5.5 (5.2)\\
\textbf{\%} &8.3 (8.0)&10.4 (9.3)&9.3 (9.2)&8.5 (7.4)&14.3 (13.4)&10.2 (10.1)\\
        \bottomrule
\end{tabular}
\end{small}
\caption{Average (standard deviation) error in step length in cm and \% of Zeno Walkway step length (\%).}
\label{tab:steplenerror}
\end{adjustwidth}
\end{table}

\subsection{Intra-Session Reliability}

To evaluate the reliability of the proposed method, we conducted an intra-session test-retest reliability analysis by comparing walks from block one to those in block two. The participants underwent an equivalent number and type of walks in both blocks to ensure consistency in measured step length. The reliability assessment is conducted using a two-way random effect, absolute agreement, multiple raters/measurement intra-class correlation (ICC) measure \cite{koo2016guideline}. The measured step length between blocks is depicted in Figure~\ref{fig:intrasessionreliability}, revealing ICC(2,k)=0.83 (95\% CI 0.77 to 0.87). This indicates a strong level of reliability for the proposed approach in intra-session assessments within a clinical setting.

\subsection{Comparison to Existing Methods}

Although a direct comparison to existing radar-based step length measurement methods is not feasible, Table~\ref{tab:SOT} provides a comparison based on the reported magnitude of error. The comparison utilizes the average from 131 Control (normal) walks by frail older adults.

The errors reported in this study are 2 to 4 times larger in magnitude when compared to existing reported figures. However, it's crucial to acknowledge the specific focus of this work on step length measurement for frail older adults. Unlike healthy subjects with a consistent gait cycle, frail older adults exhibit variability due to factors such as health conditions. This is underscored by instances where some participants were unable to complete all 20 walks due to frailty reasons. Additionally, the distance traveled is shorter, and the walks include acceleration and deceleration components, factors expected in a home setting but not accounted for in previous works.

\begin{table}
\begin{adjustwidth}{-1.5in}{0cm}
\renewcommand{\arraystretch}{1.2} 
    \centering
    \begin{small}
    \begin{tabular}{@{}lcccccc@{}}
         \toprule
         \textbf{Method} & \textbf{Avg. Error (cm)} & \textbf{total \# walks} & \textbf{\# Participants} & \textbf{Distance (m)} & \textbf{Type}  \\
         \midrule
         \cite{essay97291} & 1.1 (0.8) & 3 & 3 & 4 & Young Fit\\
         \cite{abedi2023DielectricLense} & 2.3 & 4 & 4 & 25.2 & Young Fit \\
         \cite{abedi2022Hallway} & 2.6 (1.5) & 5 & 5 & 56 & Young Fit\\
         \cite{sahoTrunkAndToe} & 2.2 (1.4) & 100 & 10 & 10 & Young Fit\\
         \midrule
         Ours & 4.5 (4.3) & 131 & 35 & 4 & Older Frail \\
         \bottomrule
    \end{tabular}
    \end{small}
    \caption{Comparison to existing methods for normal walking in a controlled setting.}
    \label{tab:SOT}
\end{adjustwidth}
\end{table}

\section{In Home Step Length Evaluation}

The measurement of step length in the home exclusively relies on the proposed radar-based system, preventing the possibility of concurrent validity assessment. Nevertheless, following literature on in-home gait speed analysis \cite{doi:10.1126/scitranslmed.adc9669}, we address the reliability of in-home step length measurement using a week-over-week test-retest framework and establish validity by correlating it with in-clinic step length measurements.

\begin{figure}
    \begin{minipage}{0.47\textwidth}
        \includegraphics[width=\linewidth]{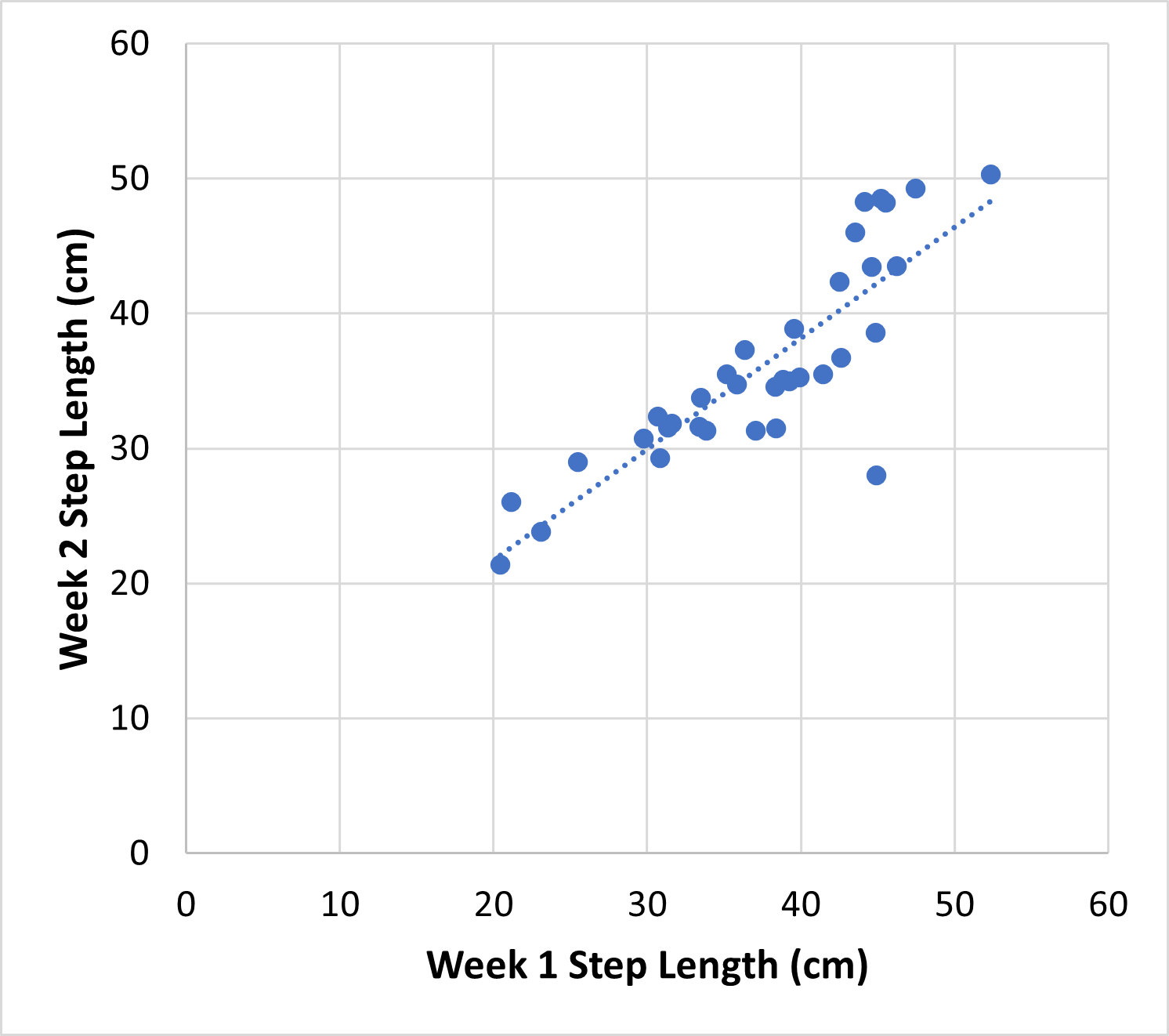}
        \caption{Average step length measurement, obtained by the proposed method, between week 1 and week 2 for each room. ICC(2,k)=0.91 (95\% CI 0.82 to 0.96), indicates excellent reliability.}
        \label{fig:reliability}
    \end{minipage}
    \hspace{0.2cm}
    \begin{minipage}{0.47\textwidth}
        \includegraphics[width=\linewidth]{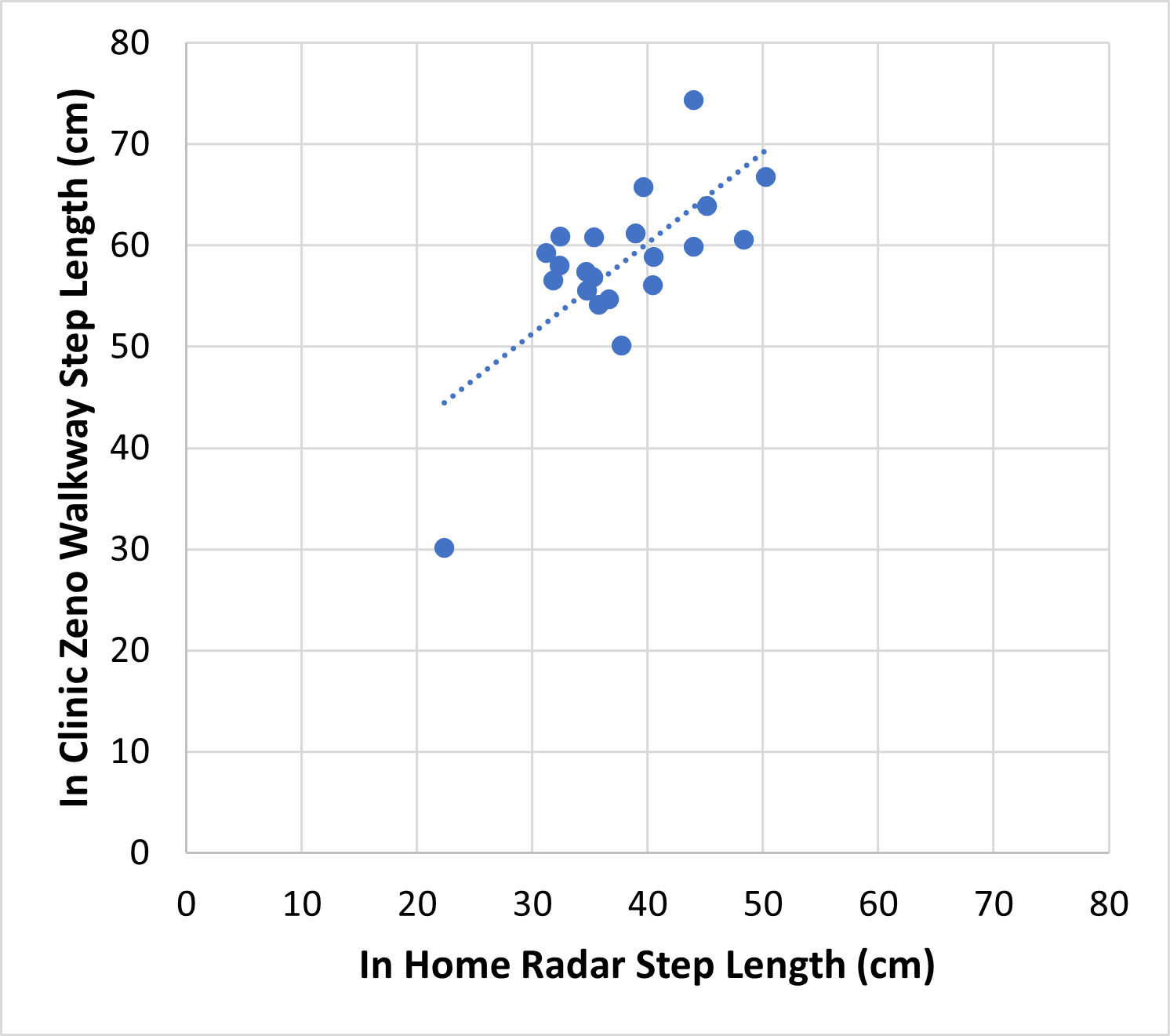}
        \caption{In home average radar step length, compared to in clinic average Control (Normal walk) step length measured by Zeno Walkway. ICC(3,k)=0.81 (95\% CI 0.53 to 0.92), indicates strong consistency.}
        \label{fig:validity}
    \end{minipage}
\end{figure}

\subsection{Reliability}

Reliability is assessed through the test-retest framework, measuring step length from week one to week two within each room of the homes. This evaluation gauges the consistency of the proposed approach for step length measurement over the two weeks, assuming no significant change in step length occurs for the participants during this period. This assumption is corroborated by an end-of-study survey where participants reported no adverse outcomes such as falls. Additionally, the use of two different devices to obtain step length in each room between week one and week two introduces a test of inter-device reliability within the test-retest framework.

Figure~\ref{fig:reliability} illustrates the average week-over-week step length measurements by the proposed approach for each room. Out of the 21 participants who set up the devices in their homes, 18 collected data over two weeks, resulting in a total of 18 participants $\times$ 3 rooms $=$ 54 rooms. However, some rooms did not have suitable tracks for measuring step lengths each week. Consequently, data from only 35 rooms across the 18 participants with reported average step lengths for the two weeks are presented in Figure~\ref{fig:reliability}.

Two-way random effect, absolute agreement, multiple raters/measurement intra-class correlation (ICC) \cite{koo2016guideline} is employed to quantify the absolute agreement between week 1 and week 2 step length measurements. Computed using the R software package v4.3.1, the ICC(2,k)=0.91 (95\% CI 0.82 to 0.96). This high ICC value indicates excellent reliability \cite{koo2016guideline} for the proposed radar-based step length measurement in a home setting.

\begin{figure}
\centering
        \includegraphics[width=0.8\linewidth]{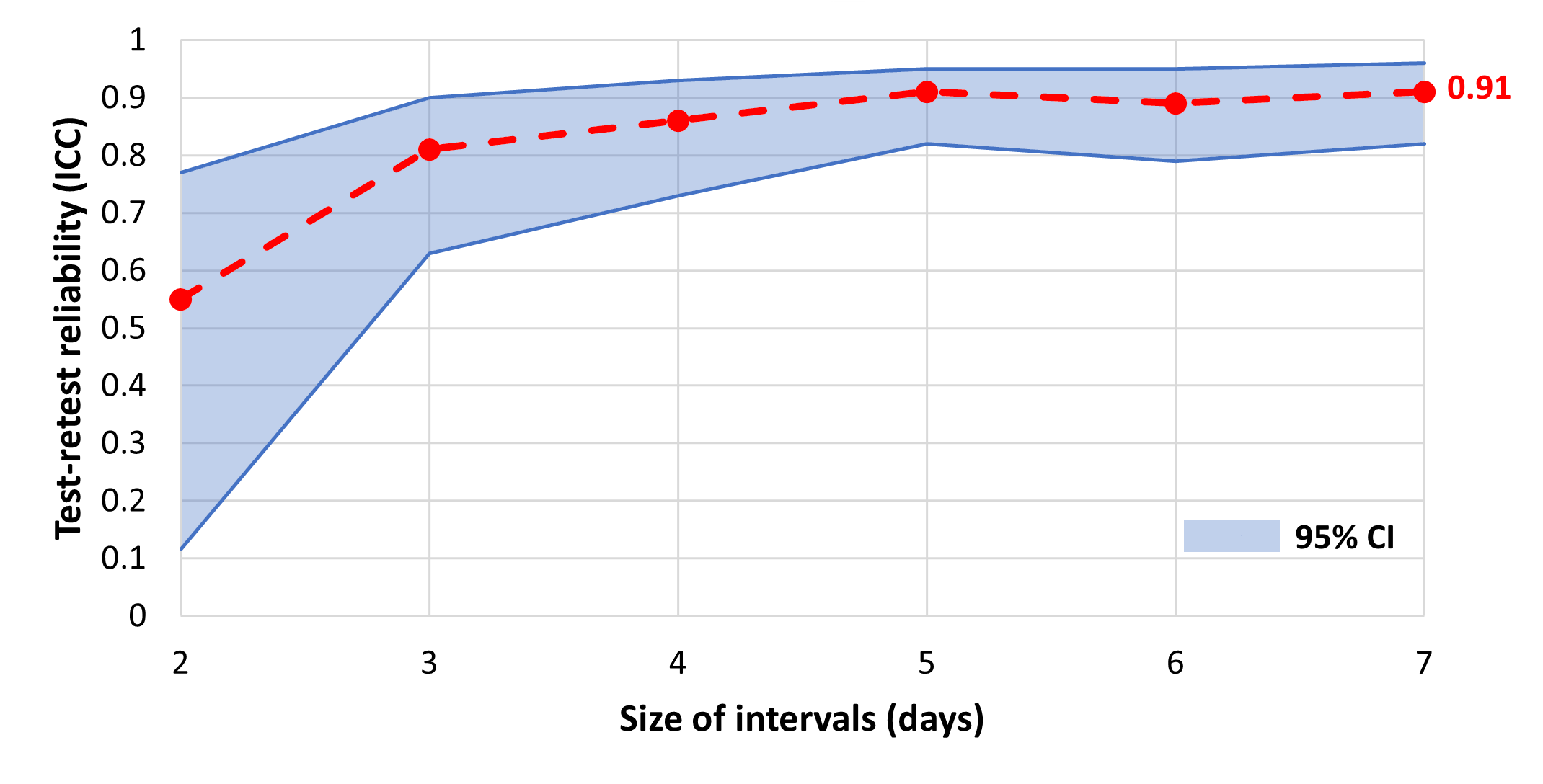}
        \caption{Test retest reliability of in home step length as a function of aggregation interval. Reliability is measured using inter-class correlation (ICC).}
        \label{fig:testretestperdays}
\end{figure}

Figure~\ref{fig:testretestperdays} plots the test retest reliability, as measured by ICC(2,k), against the step length averaging interval. The averaging interval goes from 2 days to 7 days in each week and as seen in Figure~\ref{fig:testretestperdays}, converges to an excellent reliability ($ICC(2,k)\geq0.9$) for intervals of 5 days or greater.

\subsection{Validity}
\label{sec:inhomevaldity}

Validation of step length measurement in a home setting lacks ground truth. Nevertheless, the clinical assessment of each participant incorporates the measurement of step length during normal (Control) walking. Although the step length measured during the clinical Control walk may not precisely align with the average step length measured at home, a substantial correlation is anticipated. As a result, a two-way random effects, consistency, multiple raters/measurements inter class correlation (ICC) \cite{koo2016guideline} is employed to evaluate the consistency between in home measured step length and the in clinic Control walk step length.

In Figure \ref{fig:validity}, the Control walk step length measurements in the clinic using Zeno Walkway are compared to the average in-home step length measurements based on the proposed method for all 21 participants with in-home data. Computed using R v4.3.1, ICC(3,k)=0.81 (95\% CI 0.53 to 0.92), giving good consistency between in home and in-clinic measurements \cite{koo2016guideline}. This affirms the validity of the proposed in-home step length measurement.

\begin{figure}
\begin{adjustwidth}{-1.5in}{0cm}
    \begin{subfigure}{.42\textwidth}
        \centering
        \includegraphics[width=0.97\linewidth]{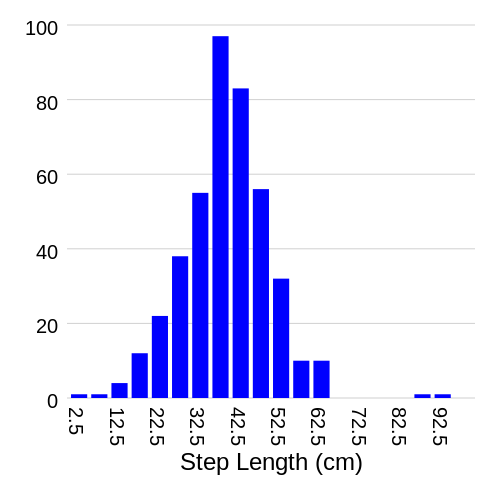}
        \caption{CHIRP001.}
        \label{fig:chirp001dist}
    \end{subfigure}
    \begin{subfigure}{.42\textwidth}
        \centering
        \includegraphics[width=0.97\linewidth]{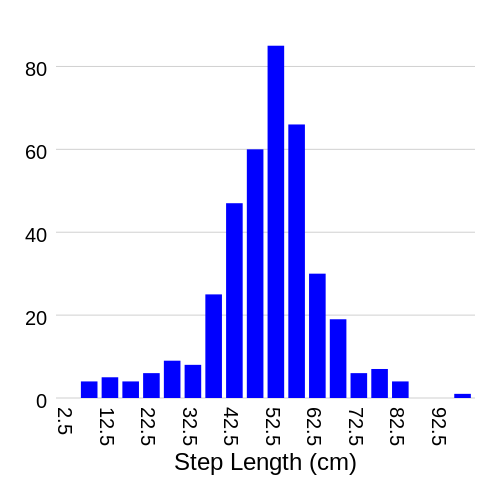}
        \caption{CHIRP019.}
        \label{fig:CHIRP019}
    \end{subfigure}
    \begin{subfigure}{.42\textwidth}
        \centering
        \includegraphics[width=0.97\linewidth]{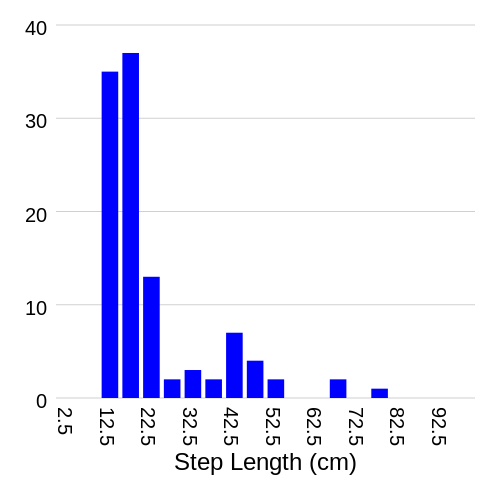}
        \caption{CHIRP022.}
        \label{fig:chirp022}
    \end{subfigure}
    \caption{Distribution of all step lengths measured in the home over the full two week data collection period.}
    \label{fig:measuredStepLengthDist}
\end{adjustwidth}
\end{figure}

While correlated with in-clinic measurement, the in-home measured step length tends to be smaller than the assessment conducted in the clinic. This phenomenon aligns with findings from other studies on gait speed, where in-home gait speed tends to be slower than that measured in a clinic setting \cite{atrsaei2021gait}.

A distinctive case worth highlighting is participant CHIRP024, who exhibited a significantly lower in-clinic step length measurement of 30cm compared to other participants. Consistently, the in-home step length measurement for CHIRP024 is notably smaller than that of the other participants, as depicted in Figure~\ref{fig:validity}. This observation further substantiates the credibility of the proposed measurement method.

Finally, Figure~\ref{fig:measuredStepLengthDist} illustrates the distribution of individual step lengths measured in the home, revealing distinct peaks corresponding to the averages identified in our previous analysis.

\section{In Home Tracks}

\begin{figure}
    \centering
    \includegraphics[width=0.8\linewidth]{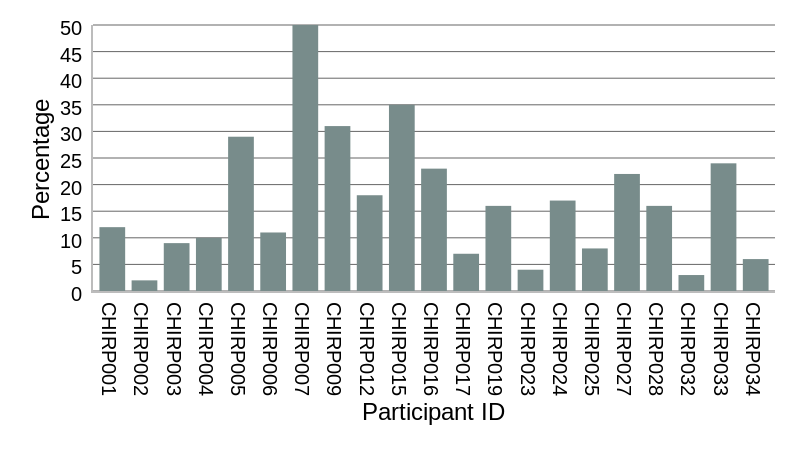}
    \caption{Percentage of valid linear track segments where step length can be measured. Valid linear track segments are linear track segments that are at least 2m in length and within $15^o$ of radar's radial axis direction.}
    \label{fig:validTracks}
\end{figure}

The proposed method relies on the assumption that, for the in-home setups selected by users, there are linear track segments that are at least $D=2$ meters long and oriented within a $\gamma=15^o$ angle of the radar's radial axis. In all 21 homes set up by participants, such tracks were identified, although their frequency varies significantly among homes. Figure~\ref{fig:validTracks} depicts the percentage of valid linear track segments, where step length can be measured. On average, valid track segments make up only a small percentage -- 17\% (12\%) -- of the total observed tracks. Additionally, three of the homes have less than 5\% valid tracks.

\begin{figure}
\begin{adjustwidth}{-1.5in}{0cm}
    \begin{subfigure}{.64\textwidth}
        \centering
        \includegraphics[width=0.98\linewidth]{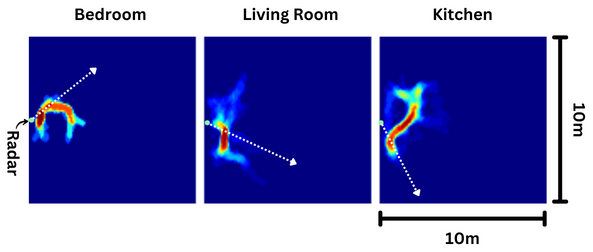}
        \caption{CHIRP002.}
        \label{fig:chirp002heatmap}
    \end{subfigure}
    \begin{subfigure}{.64\textwidth}
        \centering
        \includegraphics[width=0.98\linewidth]{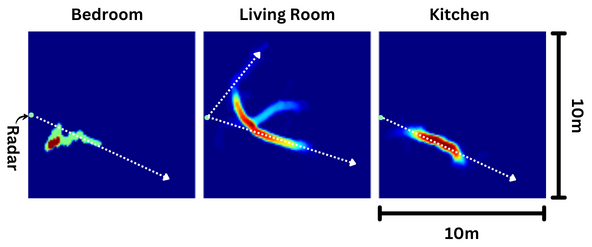}
        \caption{CHIRP007.}
        \label{fig:chirp007heatmap}
    \end{subfigure}
    \caption{Heat map of tracks in each room outlines the commonly used pathways in the home. CHIRP007 home has many commonly used pathways that lines up with radar's radial axis (illustrated in white dotted lines). CHIRP002 home's commonly used pathways do not line up with radar's radial axis.}
    \label{fig:trackHeatmaps}
\end{adjustwidth}
\end{figure}

The percentage of valid tracks is directly influenced by the home layout. For instance, consider CHIRP002, where only 2\% of the tracks are suitable for step length measurement. The heatmap displaying all tracks in the home over a day is depicted in Figure~\ref{fig:chirp002heatmap}. It is evident that the long, frequently used pathways are nearly perpendicular to the radar's radial axis. In contrast, the heatmap of tracks in CHIRP007's home is more conducive to step length measurement (Figure~\ref{fig:chirp007heatmap}). With extended pathways aligned along the radar's radial axis, nearly 50\% of the tracks are suitable for step length measurement.

\section{Conclusion}

This paper presents the first-ever assessment of radar-based step length measurement for frail older adults in both clinical and home settings, confirming the feasibility of obtaining reliable and accurate step length measurements using radar sensors. Unlike existing publications, the proposed approach for step length measurement was evaluated using 35 frail older adults in a clinical environment and 21 frail older adults in a home setting. Clinic results demonstrate that radar-based step length measurement for frail older adults is within 4.5cm of the gold standard Zeno Walkway gait analysis system and exhibits strong intra-session reliability (ICC(2,k)=0.83). In home results indicates excellent week-over-week reliability (ICC(2,k)=0.91) and a strong agreement (ICC(3,k)=0.81) between in-home and in-clinic step length measurements. Both in clinic and in home results with frail older adults validates the real-time in home step length measurement. 

Having demonstrated the feasibility of obtaining accurate and valid step length measurements in the homes of frail older adults opens up the possibility of integrating step length measurement into various approaches for assessing mobility and frailty. This includes predicting the risks of falls and hospitalizations, as well as continuously monitoring the progression of diseases such as Parkinson's.

\nolinenumbers

\bibliography{library}

\bibliographystyle{abbrv}

\end{document}